\documentclass{article}
\usepackage{ijcai24}

\usepackage{times}
\usepackage{soul}
\usepackage{url}
\usepackage[hidelinks]{hyperref}
\usepackage[utf8]{inputenc}
\usepackage[small]{caption}
\usepackage{graphicx}
\usepackage{amsmath}
\usepackage{amsthm}
\usepackage{booktabs}
\usepackage{algorithm}
\usepackage{algorithmic}
\usepackage[switch]{lineno}
\usepackage{xcolor}
\usepackage{amsmath, amssymb}
\usepackage{tikz} 
\usepackage{multirow} 
\usepackage{stfloats}

\title{Place Anything into Any Video}

\author{
    Author Name
    \affiliations
    Affiliation
    \emails
    email@example.com
}

\author{
Ziling Liu$^{1,2}$
\and
Jinyu Yang$^1$\and
Mingqi Gao$^1$\and
Feng Zheng$^{1,2}$\\
\affiliations
$^1$tapall.ai~~
$^2$Southern University of Science and Technology\\
\emails
\{ziling.liu, jinyu.yang, mingqi.gao\}@tapall.ai,
f.zheng@ieee.org
}

\begin{document}
\newcommand{\jinyu}[1]{\textcolor{red}{#1}}

\maketitle

\begin{abstract}
Controllable video editing has demonstrated remarkable potential across diverse applications, particularly in scenarios where capturing or re-capturing real-world videos is either impractical or costly. This paper introduces a novel and efficient system named \textit{Place-Anything}, which facilitates the insertion of any object into any video solely based on a picture or text description of the target object or element. The system comprises three modules: 3D generation, video reconstruction, and 3D target insertion. This integrated approach offers an efficient and effective solution for producing and editing high-quality videos by seamlessly inserting realistic objects. Through a user study, we demonstrate that our system can effortlessly place any object into any video using just a photograph of the object.
Our demo video can be found at \url{https://youtu.be/afXqgLLRnTE}.
Please also visit our project page \url{https://place-anything.github.io} to get access.
\end{abstract}

\section{Introduction}

\begin{figure}[t]
\centering
\includegraphics[width=0.9\linewidth]{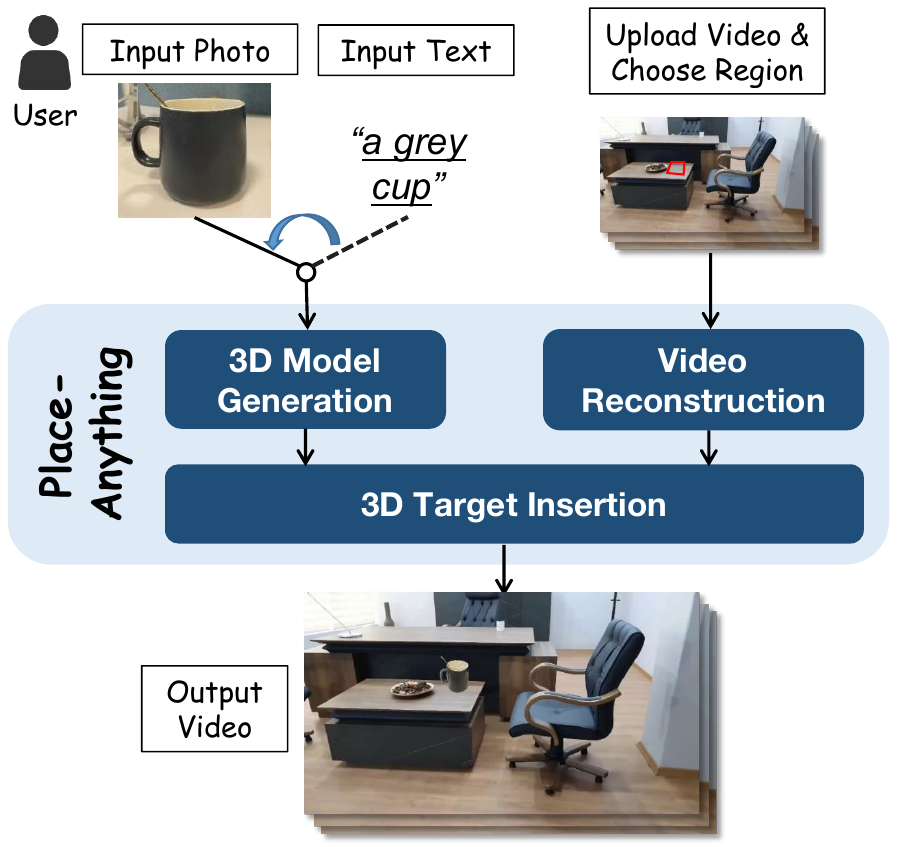}
\caption{A diagram showing how our \textit{Place-Anything} system works. The system takes source materials from users and generates required videos.}
\label{fig:user}
\end{figure}

When producing movies or commercials, acquiring video footage can often be a costly endeavor, involving significant expenses for outdoor shoots or the construction of indoor scenes.
Once filmed, modifying or editing the video content becomes expensive and time-consuming, which can only be achieved by professional post-production engineers.
For non-professionals and non-developers, inserting virtual objects into pre-existing videos to create a seamless visual effect can be an even more challenging task. 
This is primarily due to the complex operational requirements of certain post-production software and the difficulties associated with obtaining accurate 3D models. 
Consequently, there is a dire need for a user-friendly interaction solution that can simplify this process and make it more accessible. 
Moreover, such a solution can be applied to a range of applications including virtual reality, video composition, advertisement insertion, and so on.
However, achieving the goal of ``placing anything into any video'' faces multiple challenges.
For real videos, both spatial and temporal consistency are crucial, requiring inserted objects to strictly adhere to the principles of the physical world.
Nevertheless, for ease of use, the source materials provided by users should be simple and convenient, such as photos or descriptive narratives, which does not satisfy consistency well.
Moreover, for videos that need to be edited, camera parameters are mostly unknown.

As our goal is to develop user-friendly applications, we are exploring whether the ``2D $\rightarrow$ 3D $\rightarrow$ 2D'' solution can effectively address the aforementioned challenges.
To be detailed, the source materials, both photos and videos, are inherently 2D. However, the editing process occurs within a 3D space, ultimately resulting in re-created 2D videos.
Therefore, the 3D model generation and insertion become the key to such a solution.
Recently, the 3D generation techniques \cite{dreamfusion}, \cite{dreambooth3d}, \cite{liu2023zero} have gained rapid development. 
However, all methods can only generate radiance fields that are hard to use in downstream applications like 3D modification and 3D rendering. 

Even if a 3D model is obtained, traditional post-production softwares and Augmented Reality (AR) techniques, which can render the model into video, have limitations on real-world applications.
Firstly, the motion tracking tool boujou\footnote{\url{https://www.vicon.com/resources/blog/faq_tag/boujou/}} can only reconstruct the sparse point cloud from the tracking key points thus hard to localize the textureless region. 
Besides, given the results of boujou, it is still necessary to set up the camera and select a 3D plane in other 3D graphical software like 3ds Max\footnote{\url{https://www.autodesk.com.sg/products/3ds-max/overview}}, which is hard for non-professionals or non-developers.
Some AR SDKs, \textit{e.g.}, ARKit\footnote{\url{https://developer.apple.com/augmented-reality/}} in IOS and ARcore\footnote{\url{https://developers.google.com/ar}} in Android, can render the 3D model into the shooting video stream with the fusion of visual SLAM or auxiliary sensors like IMU or LiDARs. 
However, traditional SLAM pipelines depend on accurate camera intrinsic to reconstruct the scene and camera poses. 
Therefore, these AR applications gaining accurate camera intrinsic from mobile devices can not deal with arbitrary video footage offline.

To address the challenges, we introduce the \textit{Place-Anything} system, a workflow encompassing 3D model generation, video reconstruction, and 3D target insertion. As demonstrated in Figure~\ref{fig:user}, 
it is based on the 3D generation model and video self-calibration techniques, allowing users to customize their 3D assets and seamlessly integrate them into any videos, creating intriguing visual effects without prior 3D rendering knowledge. This innovative approach revolutionizes product advertisements and video post-production, allowing users to digitize tangible objects or imaginary concepts with ease. The seamless integration of reality and digital content opens new horizons for creativity and video manipulation. \textit{Place-Anything} dynamically inserts objects into diverse videos, such as advertisements and influencer content, enhancing video producers' editing capabilities and elevating viewer experiences.
Thus, \textit{Place-Anything} has at least the following advantages: \\
\noindent
\textbf{1) Versatility:} \textit{Place-Anything} boasts remarkable adaptability, allowing various objects and video scenarios. This frees users from the tedium of traditional video production and reproduction, as both tangible items and abstract concepts can be seamlessly transformed into digital assets. Moreover, it is agnostic to the source of videos, whether captured or generated, as it does not rely on specific camera parameters.\\
\noindent
\textbf{2) Interactivity:} \textit{Place-Anything} features an intuitive user interface, enabling users to create and customize videos with ease. 
With just a few clicks, users can select the desired region, adjust the scale and orientation of digital assets based on previews, and seamlessly integrate them into videos. This ensures precise and effortless video manipulation.\\
\noindent
\textbf{3) High fidelity:} 
Unlike text-to-video methods, \textit{Place-Anything} is committed to delivering high-fidelity results. Our system generates 3D meshes from photos or text descriptions, ensuring that the digital representations of objects maintain their original quality and details. This fidelity extends to the seamless integration of objects into pre-existing videos, resulting in a natural and realistic blend of content.
This level of controllability ensures that users can achieve their desired outcomes with precision while maintaining the authenticity and integrity of the original video content.

In summary, \textit{Place-Anything} offers a comprehensive solution for video manipulation, combining versatility, interactivity, and high fidelity to deliver exceptional results for users. Whether you're a professional video producer or a casual user, our system enables you to seamlessly blend reality and digital content, unlocking a world of creative possibilities.

\begin{figure*}[t]
\centering
\includegraphics[width=0.85\linewidth]{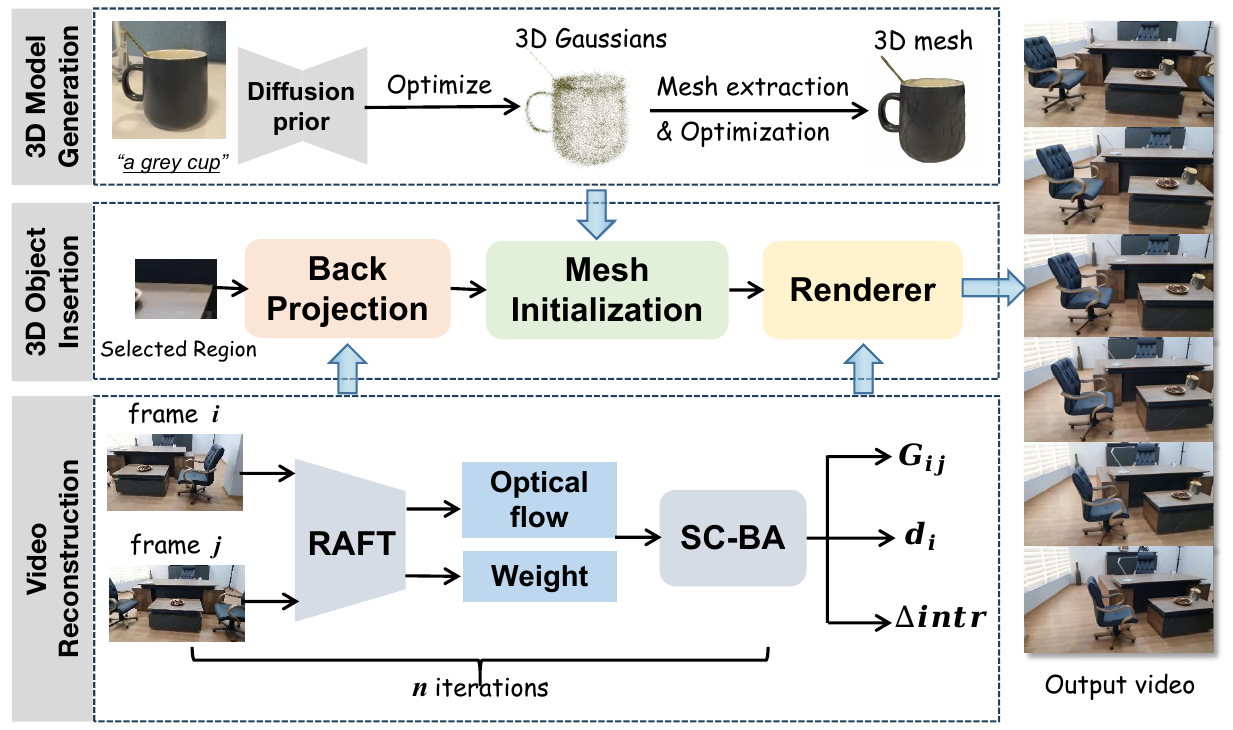}
\caption{Overview of the pipeline of our \textit{Place-Anything} system.}
\label{fig:pipeline}
\end{figure*}

\section{System Architecture}

\textbf{Overview.} Figure~\ref{fig:pipeline} gives an overview of the proposed \textit{Place-Anything}.
To begin, users simply need to capture a photo of a physical object or provide a brief description of a virtual one. Within minutes, our system generates a corresponding 3D mesh, accurately representing the object's form and details. 
Next, users upload a video, and our advanced system promptly estimates the camera's intrinsic parameters. 
It then reconstructs the camera poses and dense depth maps for every frame, ensuring precise alignment and integration of 3D content.
At this point, users can select a region on the first frame using points or boxes, specifying where they want to place the 3D object. 
They can fine-tune the model's scale and orientation based on preview results, ensuring it fits seamlessly into the video's context. Once these steps are completed, our efficient renderer parallelly generates the corresponding multi-view images of 3D objects for each frame, seamlessly blending them into the video's flow. 
The beauty of our system lies in its simplicity and accessibility. Unlike traditional 3D construction and rendering pipelines, all operations and interactions occur in 2D, making it easy for anyone to use. 
Regardless of skill level, \textit{Place-Anything} enables users to create immersive videos effortlessly with just a few clicks.
In the following, we will introduce each module.

\begin{figure*}[t]
\centering
\includegraphics[width=\linewidth]{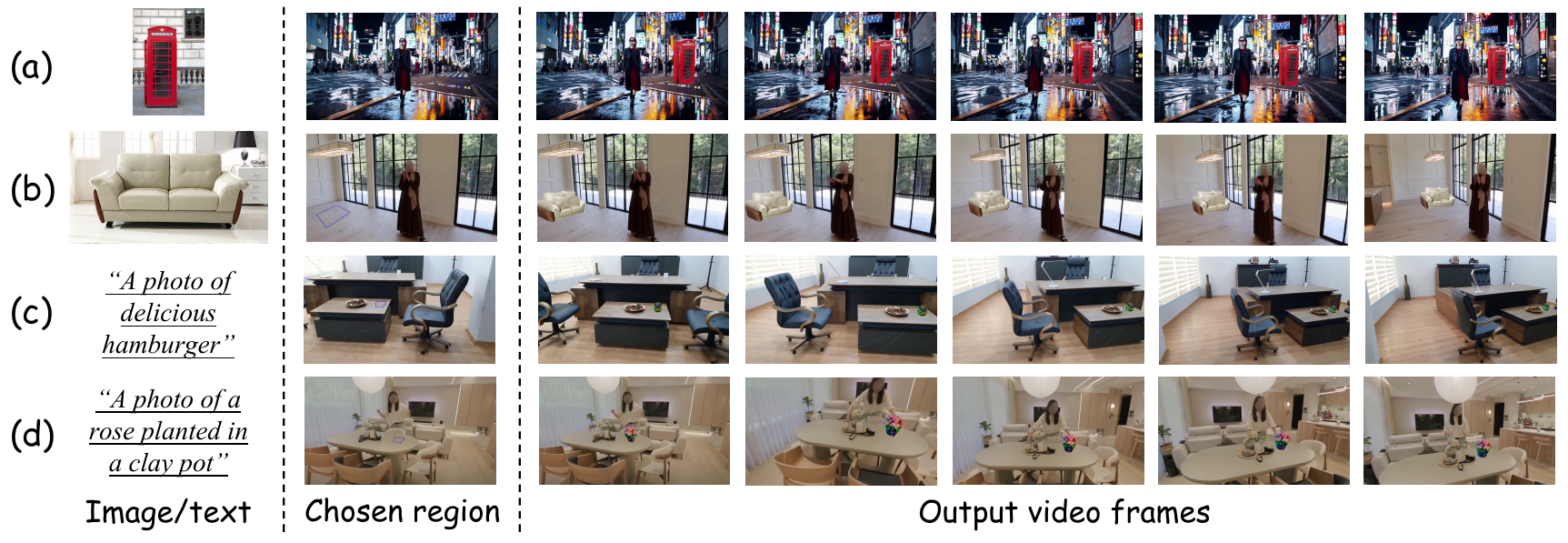}
\caption{Visualized examples generated by \textit{Place-Anything}. } 
\label{fig:results} 
\end{figure*}

\noindent
\textbf{3D Model Generation.}
With the given photo or text description of an object, the first step is to generate its 3D model.
While existing methods (\cite{dreamfusion,dreambooth3d,liu2023zero}) can generate radiance fields from images or text, they suffer from limited generation speed due to volume rendering and pixel-by-pixel optimization. These fields, unlike meshes, are challenging to scale, translate, or rotate, making them unsuitable for subsequent rendering. In contrast, the advanced 3D framework \cite{tang2023dreamgaussian} offers easy-to-optimize 3D representations and efficient mesh extraction. We adopt it as our 3D generator, where users upload an object's description or image. We train a 3D Gaussian \cite{3dGS} supervised by diffusion, then recover mesh geometry by dividing Gaussians into uniform grids, calculating weighted opacity, and using Marching Cubes for mesh extraction. Finally, we bake the colored texture map by back-projecting rendered RGB onto the mesh surface.

\noindent
\textbf{Video Reconstruction.}
This module aims to estimate camera intrinsics, reconstruct camera poses, and generate dense depth maps for input videos. 
Previous SLAM systems~\cite{campos2021orb,engel2017direct}, relying on key point tracking and epipolar geometry, struggle with weak or repetitive textures, making it challenging to place virtual objects on smooth surfaces. 
They also cannot handle videos with unknown cameras. 
To address these issues, we utilize \cite{droid-calib}. 
We first predict weighted optical flow between frames using RAFT~\cite{teed2020raft}. 
Then a set of key frames will be filtered out based on sufficient optical flow displacement and this set of key frames will be continuously updated online every time a new frame enters in a sliding window manner. 
In a single window, a self-calibrating weighted bundle adjustment (SC-BA) \cite{droid-calib} which minimizes the reprojection errors can estimate the relative camera poses $\mathbf{G}_{ij}$, pixels depth $\mathbf{d}_{i}$ and intrinsic update $\Delta \mathbf{intr}$ through differentiable Gauss-Newton steps. 
Dense depth maps from bundle adjustment, aided by optical flow, locate any area in a video frame.

\noindent
\textbf{3D Target Insertion.}
The obtained 3D model and reconstructed video are combined to create the new video. 
Users can select the region for placement in the first frame via clicks or bounding boxes.
Using pixel coordinates, depth values, and camera intrinsics/extrinsics, we back-project the chosen region to 3D points:
$[{x}_{i}, {y}_{i}, {z}_{i}] = \mathbf{G}_{i} \otimes \boldsymbol{\pi}^{-1}\left(\mathbf{u}_{i}, \mathbf{z}_{i}, \boldsymbol{\theta}\right)$,
where $\boldsymbol{\pi}^{-1}:\mathbb{R}^2 \times \mathbb{R} \rightarrow \mathbb{R}^3$ denotes the inverse projection, $\mathbf{G}_{i}$ denotes the camera to world projection matrix, $\mathbf{u}_{i}$ denotes the pixel coordinate, $\mathbf{z}_{i}$ is the depth value of the pixel, and $\boldsymbol{\theta}$ is the camera intrinsic.
We then estimate the plane function using RANSAC on the projected 3D points. Constraints on the normal vector ensure the 3D model is placed upright. 
Specifically, given the normal vector as $\overrightarrow{\mathit{N}}$, camera position as $c$, and a point in the plane as $p$, it satisfies that $\overrightarrow{\mathit{N}} \cdot\overrightarrow{\mathit{c}\mathit{p}} > 0$.
After rotating the 3D model to align its vertical y-axis with the normal vector, we adjust its scale based on the target area's extension range along the x- and y-axis.
Finally, we render the multi-view images in parallel using pytorch3d~\cite{pytorch3d} and composite the rendering results with background video.

\section{Demostrations}

Figure~\ref{fig:results} presents example videos generated by \textit{Place-Anything}, showcasing the advanced capabilities of our system. Note that the source video of (a) is generated by SORA~\cite{sora} while others are downloaded from YouTube.
Firstly, the generated 3D model maintains strong visual coherence with the input reference images, thanks to the robust diffusion prior. 
Secondly, our system efficiently tracks and identifies textureless regions by leveraging optical flow to establish pixel correspondence between adjacent frames. Consequently, 3D models can be seamlessly integrated onto smooth surfaces like desktops or floors, regardless of the absence of corners or prominent textures. 
Furthermore, our model has successfully inferred the video's camera intrinsics and pose, relying solely on the harmonious and stable rendering of each frame. 
Lastly, mesh initialization in the third module guarantees that the inserted 3D model's scale matches the selected area perfectly.
More cases can be found on both the demo video\footnote{\url{https://youtu.be/afXqgLLRnTE}} and our project page\footnote{\url{https://place-anything.github.io/}}.

\section{Applications}
\textit{Place-Anything} is also adaptable to various video applications. We here show some potential applications.

\textbf{Product Advertisement and Marketing:} Brands can use \textit{Place-Anything} to create personalized customized advertisements by inserting 3D models of their products into realistic videos. This can showcase new products in a realistic environment and can be updated at any time.

\textbf{Influencer Marketing:} Social media influencers can integrate \textit{Place-Anything} into their content creation process. By inserting personalized 3D models or props into their videos, they can create more engaging and unique content that stands out from the crowd.

\textbf{VR and AR Applications:} \textit{Place-Anything} can be integrated into VR and AR applications to create more immersive experiences. Users can place 3D models of objects or environments into their VR or AR worlds, enhancing the overall experience.

\textbf{Video Editing and Post-Production:} Video editors and producers can use \textit{Place-Anything} to enhance their videos with dynamic 3D insertions. This can include adding special effects, creating composite shots, or inserting 3D elements into existing footage.

\textbf{Entertainment Industry:} The entertainment industry can benefit from \textit{Place-Anything} by using it to create realistic and immersive scenarios in movies, television shows, or games. By inserting 3D models into backgrounds or scenes, producers can create believable and engaging visuals.

\textbf{Real Estate and Architecture:} Real estate agents and architects can use \textit{Place-Anything} to showcase properties or designs in a more interactive and engaging way. By inserting 3D models of furniture or fixtures into videos, they can provide potential buyers or clients with a better understanding of the space.

These are just a few examples of how \textit{Place-Anything} can be applied in various industries and scenarios. The system's flexibility and ease of use make it a powerful tool for creative expression and video manipulation.

\section{Conclusion}
In this paper, we present a novel system, called \textit{Place-Anything}, which efficiently completes the entire process from 3D model production to embedding 3D models into existing videos. The simple 3D model production methods and interactive approaches make it possible for anyone to effortlessly integrate objects from their imagination or immediate environment into the creative process of any pre-existing video.


\bibliographystyle{named}

\end{document}